\documentclass[a4paper,fleqn,6pt]{cas-dc}

\usepackage{setspace}
\usepackage{tcolorbox}
\usepackage{soul}
\usepackage{xcolor}
\usepackage{lineno}
\usepackage{xcolor}

\usepackage[numbers,square,comma,sort&compress]{natbib}
\usepackage{multicol}
\usepackage{color}
\usepackage{xcolor}
\usepackage{hyperref}
\hypersetup{
    colorlinks=true,
    linkcolor=blue, 
    citecolor=blue,
    urlcolor=blue
}

\makeatletter
\def\tagform@#1{\maketag@@@{(\ignorespaces#1\unskip\@@italiccorr)}}
\renewcommand{\eqref}[1]{\textcolor{blue}{\textup{\textrm{Eq }{(\ref{#1})}}}}
\makeatother

\usepackage[doipre={doi:~}]{uri}
\usepackage{indentfirst}
\usepackage{setspace}
\usepackage{graphicx}
\usepackage{ragged2e}
\usepackage[labelfont=rm,textfont={rm,it}]{caption}
\def\tsc#1{\csdef{#1}{\textsc{\lowercase{#1}}\xspace}}
\tsc{WGM}
\tsc{QE}
\tsc{EP}
\tsc{PMS}
\tsc{BEC}
\tsc{DE}
\makeatletter
\renewcommand{\fnum@figure}{Fig. \thefigure.\@gobble}
\makeatother

\begin{document}
\setstretch{2} 
\sethlcolor{yellow}
\let\WriteBookmarks\relax
\def\floatpagepagefraction{1}
\def\textpagefraction{.001}
\let\printorcid\relax

\title [mode = title]{RS-MOCO: A deep learning-based topology-preserving image registration method for cardiac T1 mapping}                      

\author[1,2]{Chiyi Huang}
\fnmark[1] 
\fntext[1]{These authors contributed equally to this work.}

\credit{Conceptualization of this study, Methodology, Software}

\affiliation[1]{organization={Paul C. Lauterbur Research Center for Biomedical lmaging, Shenzhen Institutes of Advanced Technology. Chinese Academy of Sciences},
    % addressline={Radarweg 29},
    city={Guangdong},
    % citysep={}, % Uncomment if no comma needed between city and postcode
    postcode={518055}, 
    country={China}}

\affiliation[2]{organization={University of Chinese Academy of Sciences},
    % addressline={Radarweg 29},
    city={Beijing},
    % citysep={}, % Uncomment if no comma needed between city and postcode
    postcode={100049}, 
    country={China}}
    
\author[3]{Longwei Sun}[type=editor,
                        auid=000,bioid=1,
                        % role=Researcher,
                        % orcid=0000-0001-7511-2910
                        ]
\fnmark[1] 

\affiliation[3]{organization={Department of Radiology, Shenzhen Children's Hospital},
    % addressline={Radarweg 29},
    city={Guangdong},
    % citysep={}, % Uncomment if no comma needed between city and postcode
    postcode={518034}, 
    country={China}}

\author[1]{Dong Liang}[type=editor,
                        auid=000,bioid=1,
                        % role=Researcher,
                        % orcid=0000-0001-7511-2910
                        ]                       

\author[1]{Haifeng Wang}[type=editor,
                        auid=000,bioid=1,
                        % role=Researcher,
                        % orcid=0000-0001-7511-2910
                        ]

\author[3]{Hongwu Zeng}[type=editor,
                        auid=000,bioid=1,
                        % role=Researcher,
                        % orcid=0000-0001-7511-2910
                        ]
\cormark[1]
\ead{homerzeng@126.com}

\author[1]{Yanjie Zhu}[type=editor,
                        auid=000,bioid=1,
                        % role=Researcher,
                        % orcid=0000-0001-7511-2910
                        ]
\cormark[1]
\ead{yj.zhu@siat.ac.cn}

\cortext[cor1]{Corresponding author.}
% \cortext[cor2]{Principal corresponding author}

\begin{abstract}
Cardiac T1 mapping can evaluate various clinical symptoms of myocardial tissue. However, there is currently a lack of effective, robust, and efficient methods for motion correction in cardiac T1 mapping. In this paper, we propose a deep learning-based and topology-preserving image registration framework for motion correction in cardiac T1 mapping. Notably, our proposed implicit consistency constraint dubbed BLOC, to some extent preserves the image topology in registration by bidirectional consistency constraint and local anti-folding constraint. To address the contrast variation issue, we introduce a weighted image similarity metric for multimodal registration of cardiac T1-weighted images. Besides, a semi-supervised myocardium segmentation network and a dual-domain attention module are integrated into the framework to further improve the performance of the registration. Numerous comparative experiments, as well as ablation studies, demonstrated the effectiveness and high robustness of our method. The results also indicate that the proposed weighted image similarity metric, specifically crafted for our network, contributes a lot to the enhancement of the motion correction efficacy, while the bidirectional consistency constraint combined with the local anti-folding constraint ensures a more desirable topology-preserving registration mapping.
\end{abstract}

{\setstretch{1}
}

\maketitle
\setstretch{1}
\section{Introduction}
\begin{small}
% \begin{linenumbers}
Magnetic resonance cardiac T1 mapping is a useful technique that quantitatively assesses the longitudinal relaxation time (T1) of myocardial region. The obtained myocardial T1 map can be used as a pathophysiological biomarker in various cardiac diseases, such as diffuse myocardial fibrosis \cite{Equilibrium}, myocardial infarction \cite{Myocardial}, and cardiac amyloidosis \cite{Cardiovascular}. Typical cardiac T1 mapping techniques include MOLLI \cite{Modified}, SASHA \cite{SASHA}, and STONE \cite{STONE}. These techniques acquire several cardiac T1-weighted (T1w) images with different inversion times (TIs) using electrocardiogram gating. Subsequently, pixel-wise curve fitting is applied to multiple T1w images to obtain myocardial T1 maps. However, due to the non-rigid motion of the heart, misalignment in myocardial position and shape between different T1w images can result in significant motion artifacts in the fitted T1 maps, leading to errors in myocardial T1 quantification.

To mitigate this impact, image registration is commonly performed to align the myocardial region of multiple T1w images. Nevertheless, there exist substantial and non-uniform contrast variations between T1w images with different TIs, making myocardial registration challenging. As shown in \textbf{\footnotesize\autoref{Fig. 1}}, the signal intensities of the myocardium, blood pool, and fat vary with TIs and experience signal nulling at different inversion times \cite{MotionXue}. Distinguishing myocardium from the blood pool is difficult when they manifest similar signal intensities (shown in \textbf{\footnotesize\autoref{Fig. 1}\textcolor{blue}{(3)}}), resulting in registration errors in these regions. In other words, these multi-modality image characteristics make traditional intensity-based registration methods inadequate in effectively solving the motion correction issue in cardiac T1 mapping. 

\begin{figure}
	\justify
        \includegraphics[scale=.55]{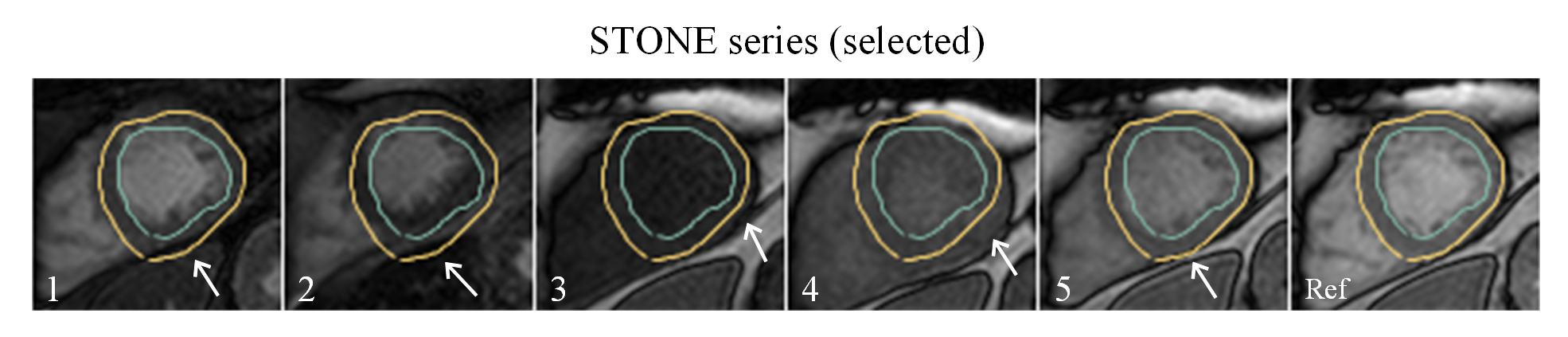}
	\caption{\textrm{The figure displays $6$ example T1w images selected from a STONE series. The endo and epi contours of the myocardium were manually delineated on the reference image (Ref) and applied to the other $5$ images to demonstrate the misalignment of each frame with respect to the reference image, as indicated by white arrows.}}
 \label{Fig. 1}
\end{figure}

To address this issue, Roujol et al. \cite{Adaptive} simultaneously estimated motion fields and intensity variations using adaptive affine registration and refined deformable registration. However, specific parameter settings may not meet the correction requirements for all motion artifact scenarios. Huizinga et al. \cite{PCA-based} and Tao et al. \cite{Robust} employed a groupwise registration approach based on principal component analysis to address motion correction in cardiac T1 mapping, avoiding the significant contrast variation issues faced by pairwise registration. Nevertheless, their approach did not consider topology preservation in deformation, which may lead to undesirable warping. Zhang et al. \cite{Cardiac} performed pairwise multimodal registration of cardiac T1w images using normalized gradient fields as the similarity metric and applied additional regularization constraints to prevent erroneous deformations. El-Rewaidy et al. \cite{Nonrigid} proposed a non-rigid registration framework based on an active shape model for motion correction of cardiac T1 mapping. This method first extracts the myocardium contours using an active shape model and then performs contour-based registration on the myocardium. However, this method may fail in image registration due to the incorrect extraction results of myocardium contours. In recent years, deep learning-based registration methods have shown higher efficiency and robustness compared to traditional registration algorithms and have gradually approached or even surpassed them in accuracy \cite{Deep}. Gonzales et al. \cite{MOCOnet} proposed a groupwise motion correction network based on U-net, which predicts relative displacement vector fields between all T1w images in a ShMOLLI series, avoiding the challenges posed by large contrast variations, but is unable to correct severe motion artifacts. Li et al. \cite{MotionLi} used sparse coding to achieve contrast separation and obtain T1w images with similar contrast for intensity-based registration. Inspired by disentangled representation learning, Yang et al. \cite{DisQ} decomposed cardiac qMRI images into anatomical representations and contrast representations to facilitate intensity-based registration.

In summary, both conventional and deep learning-based methods either employ direct multi-modality image registration of cardiac T1w images or execute intensity-based registration through utilizing groupwise registration or synthesizing T1w images with little contrast differences. Although the aforementioned registration-based motion correction approaches have achieved promising results, most of them do not consider the topological preservation properties in the ideal registration mapping. Topological preservation means that the mapping between the moved image and the moving image should be continuous, invertible, and bijective. In terms of image appearance, this implies that the deformed image (moved) should exhibit consistent local topological structure and spatial correspondence with the original image (moving), and should not contain regions with discontinuities, folds, or abnormal distortions. It is noteworthy that topology-preserving registration refers to the process of aligning the anatomical structures within the moving image with their corresponding anatomical structures in the fixed image, in terms of both position and shape, without disrupting the local topology of the anatomical structures in the moving image. Furthermore, unique anatomical structures present only in the fixed image will not result in the emergence of equivalent structures in the deformed image (moved image), as these anatomical structures do not exist in the moving image. Additionally, studies have shown that using segmentation labels for supervised or semi-supervised registration can improve registration performance \cite{VoxelMorph}. Attention modules incorporated into the backbone of convolutional neural networks also make efforts to the enhancement of network efficiency\cite{Cbam}.  

Therefore, this paper proposes a deep learning-based, bidirectional consistent, and local neighborhood-consistent image registration method for motion correction in cardiac T1 mapping. The main idea is to utilize a deformable registration network with an implicit regularization constraint (dubbed BLOC), which consists of a bidirectional consistency constraint and a local anti-folding constraint. BLOC constraint achieves topology preservation by employing the local anti-folding constraint to the predicted bidirectional deformation fields, and then inverting the two deformation fields to add consistency constraints between the original input image and its re-deformed image predicted by the network using the inversed deformation field on both sides. The proposed BLOC constraint ensures the invertibility and bijection of the registration mapping, thereby reducing abnormal deformations such as discontinuities, folds, or excessive distortions. To address the contrast variation issue, a weighted image similarity metric is designed for multi-modality image registration. Additionally, we integrate the dual-domain attention modules into the backbone network and leverage an additional semi-supervised myocardium segmentation network to perform weak supervision for the registration network, thereby improving registration accuracy. The main contributions of this paper are as follows: 

\begin{enumerate}
\itemsep=0pt
\item  A deformable registration network with a dual-domain attention module and a weighted image similarity metric is proposed to effectively learn the multi-modality image registration mapping of cardiac T1w images under significant and non-uniform contrast variations;
\item  A BLOC constraint is utilized in the deformable registration network to ensure the topology preservation in image registration, bringing about more desirable non-rigid registration outcomes;
% \begin{linenumbers}
\item  An additional semi-supervised myocardium segmentation network is incorporated to provide weakly-supervised pseudo-labels of myocardium segmentation to assist the deformable registration network.
\end{enumerate} 
\end{small}

\section{Related work}
\begin{small}
VoxelMorph \cite{AnBalakrishnan, VoxelMorph} is a deep learning-based medical image registration framework that is widely used and has established a foundation for subsequent research. For topology preservation in medical image registration, many studies \cite{UnsupervisedDalca, UnsupervisedKim, CycleMorph, Cycle-consistent, Fast, Symmetric, Inverse-consistent, DeepChen} have been proposed. For example, ICNet \cite{Inverse-consistent} ensures the inherent inverse-consistent property for diffeomorphic mapping through the application of inverse-consistent and anti-folding constraints on the displacement vector field. CycleMorph \cite{CycleMorph} applied the cycle consistency constraint between the original moving image and its re-deformed image using two registration networks. SYMnet \cite{Fast} simultaneously estimates two symmetric deformation fields and employs a combination of forward mapping and inverse backward mapping to guarantee the reversibility and inverse consistency of the registration mapping. Additionally, research has shown that deep learning-based image registration and segmentation may improve both performances by mutual supervised learning \cite{Contrastive, DeepAtlas, Anatomy-guided}. DeepAtlas \cite{DeepAtlas} first proposed a joint learning of two deep neural networks for image registration and segmentation respectively, achieving significant improvements in segmentation and registration accuracy.

The attention module is a special neural network that can be grafted into different positions of the backbone network. It applies different weights to different parts of the network in order to focus on important feature spaces or channels, thus capturing those most critical features from the input images. In recent years, the attention module has rapidly evolved, encompassing spatial attention, channel attention, and self-attention, among others. The SE-Net introduced by Hu et al. \cite{Squeeze-and-excitation} proposed a channel attention module that explicitly models the interdependencies between channels to recalibrate channel responses and focus on important channel information. Woo et al. \cite{Cbam} proposed an attention module that combines feature channels and spatial information, simultaneously attending to important channel and spatial information. Furthermore, there is also the cross-modal attention module \cite{Cross-modal} designed for multimodal medical image registration, which establishes spatial correspondences between features from different modalities for capturing information more superiorly.
\end{small}

\section{Method} 
\begin{figure*} % [width=\textwidth] 和 *表示两栏
	\centering
	   \includegraphics[width=0.8\textwidth]{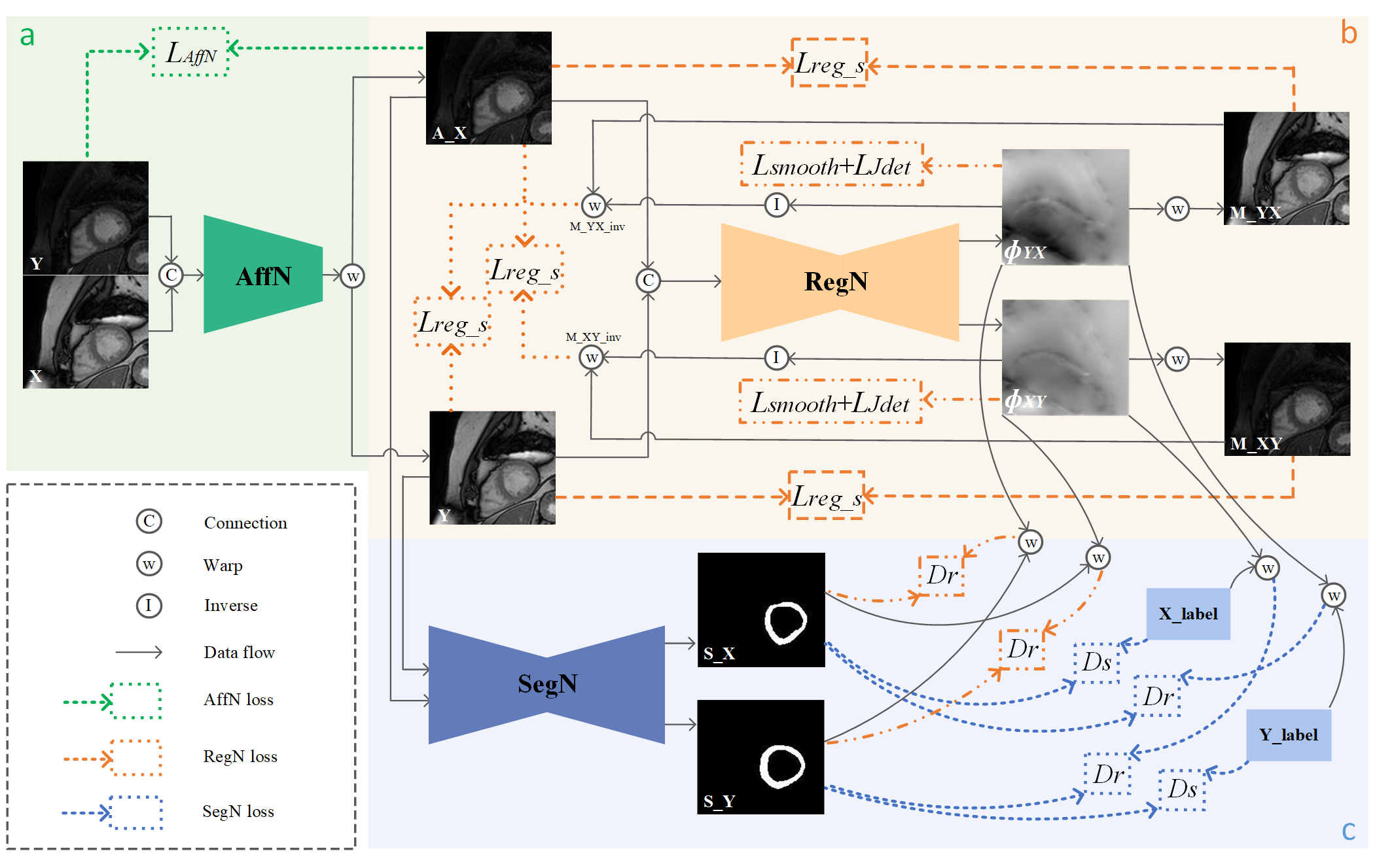}
    \caption{\textrm{The overall workflow diagram of RS-MOCO.}}
    % \textrm{\footnotesize{\textbf{Fig. 2.} The overall workflow diagram of RS-MOCO.}}
    \label{Fig. 2} 
\end{figure*}

\begin{small}
The framework of the proposed RS-MOCO method consists of three components, as illustrated in \textbf{\footnotesize\autoref{Fig. 2} \textcolor{blue}{(a-c)}}. The three components are: (1) an affine registration network (AffN) based on CNN(Convolutional Neural Network) (areas (a) in light green) for pre-training, achieving initial linear affine registration, (2) a weakly-supervised and topology-preserving deformable registration network (RegN) based on U-net architecture (areas (b) in light yellow), and (3) a semi-supervised myocardium segmentation network (SegN) for providing pseudo-labels of myocardium segmentation for RegN, also based on U-net architecture (areas (c) in light blue). The input of the network is the pair of the moving image X and the fixed image Y. $\mathrm{A}\textunderscore\mathrm{X}$ is the output of the affine registration network, which registers X to Y, while $\mathrm{M}\textunderscore\mathrm{YX}$ and $\mathrm{M}\textunderscore\mathrm{XY}$ are the deformable registration results that register Y to $\mathrm{A}\textunderscore\mathrm{X}$ and $\mathrm{A}\textunderscore\mathrm{X}$ to Y. 
\end{small}

\subsection*{\textnormal{\textit{\fontsize{10}{20}\selectfont 3.1. Affine registration network (AffN)}}}
\begin{small}
Our method begins by utilizing a CNN-based affine registration network, AffN, to rectify linear misalignment among multiple T1w images of cardiac T1 mapping. Pre-alignment using affine registration may help address the severe misalignment which is hard to deal with using deformable registration \cite{DeepStrittmatter, AHuang, DeepASDM}. Then, the six transformation parameters predicted by the proposed AffN can be applied to the spatial transformer network, STN \cite{Spatial}, to execute the corresponding translation, scaling, and rotation on input images. STN is a differentiable module providing smooth deformed feature maps by employing bilinear interpolation. It can be inserted at any position within a convolutional network architecture.

The AffN can be parameterized as the following optimization problem: 
\begin{equation}
\hat{A_{\theta}} = \text{argmin}\mathcal{L}_{dissim}(Y, \mathcal{F}_{A}(X, A_{\theta}))\label{eq:1}
\tag{1}
\end{equation}

Where represents the AffN, registering the moving image X to the fixed image Y with parameters $A_{\theta}$ illustrated in \textbf{\footnotesize\eqref{eq:2}}, and $\mathcal{L}_{dissim}$ denotes the image difference between Y and the moved image $\mathcal{F}_{A}(X, A_{\theta})$ after the affine transformation.

\begin{equation}
\textit{A}_{\theta} = \begin{bmatrix}
\mathbf{\theta}_{11} & \mathbf{\theta}_{12} & \mathbf{\theta}_{13} \\
\mathbf{\theta}_{21} & \mathbf{\theta}_{22} & \mathbf{\theta}_{23} \\
\end{bmatrix}\label{eq:2}
\tag{2}
\end{equation}
\end{small}

\begin{small}
The detailed network architecture is shown in \textbf{\footnotesize\autoref{Fig. 3}}. The AffN consists of six downsampling modules (ADSM) and two fully connected layers. Additionally, dropout layers are added at the end of the sixth downsampling block and the first fully connected layer to alleviate overfitting.

\begin{figure}  % [width=\textwidth] 和 *表示两栏
	\centering
		\includegraphics[scale=.61]{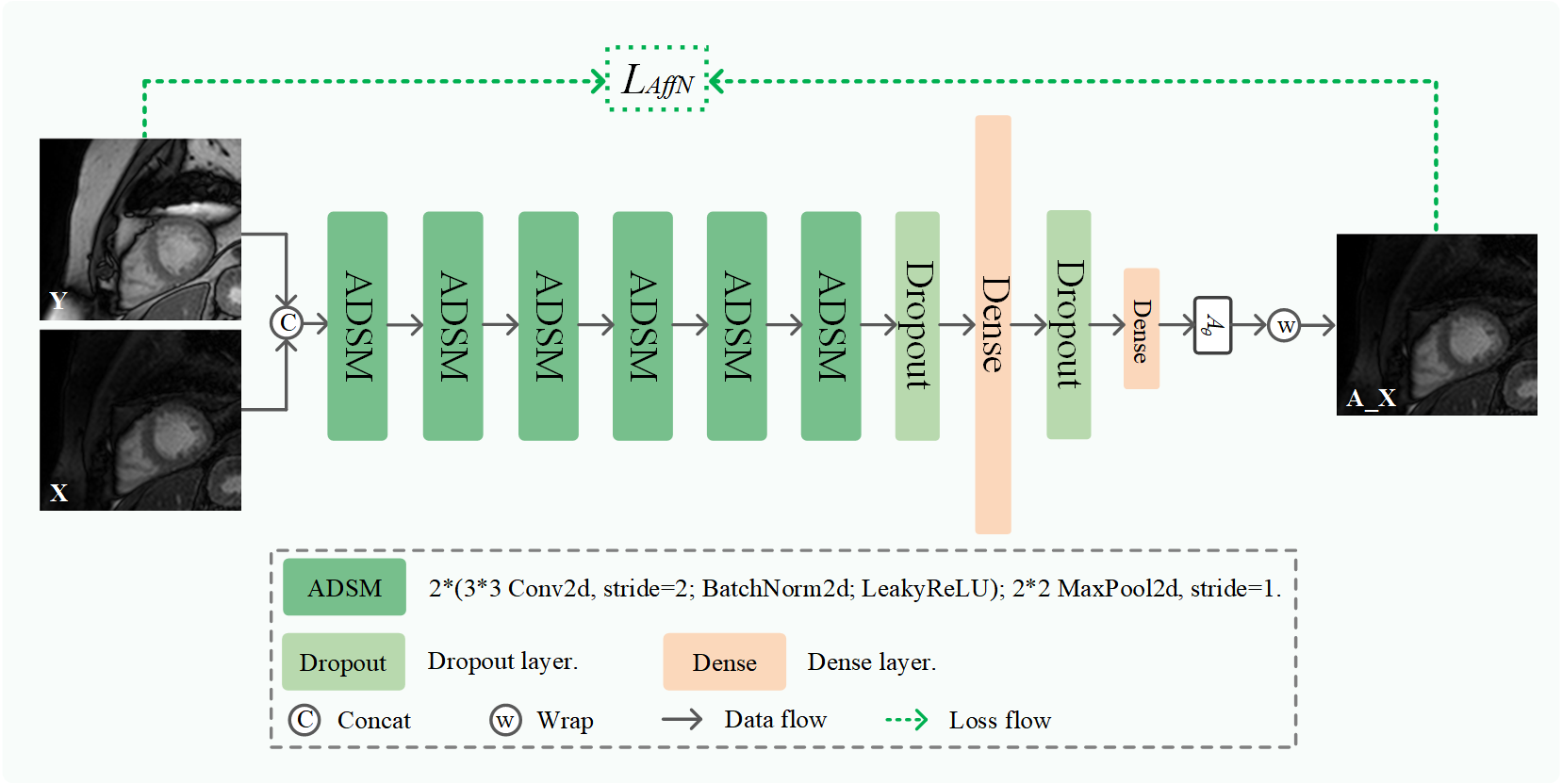}
  
	\caption{\textrm{Implementation details of the network architecture for AffN.}}
\label{Fig. 3}
\end{figure}

\end{small}

\subsection*{\textnormal{\textit{\fontsize{10}{20}\selectfont 3.2. Deformable registration network (RegN)}}}
\subsubsection*{\textnormal{\textit{\fontsize{9}{20}\selectfont 3.2.1. Baseline design}}}
\begin{small}
Generally, the training process of deep learning-based registration network training aims to minimize the difference between the original fixed image and the deformed moving image. In this study, the proposed RegN simultaneously predicts the bidirectional deformation fields from $\mathrm{A}\textunderscore\mathrm{X}$ to Y which is marked as $\mathrm{\Phi_{XY}}$ and from Y to $\mathrm{A}\textunderscore\mathrm{X}$ which is $\mathrm{\Phi_{YX}}$, as shown in \textbf{\footnotesize\autoref{Fig. 2} \textcolor{blue}{(b)}}. To achieve this, RegN minimizes the differences of the following image pairs, which are connected by yellow dashed lines in \textbf{\footnotesize\autoref{Fig. 2} \textcolor{blue}{(b)}}: direction (1): the input image $\mathrm{A}\textunderscore\mathrm{X}$ and the output image $\mathrm{M}\textunderscore\mathrm{YX}$ which is warped from Y by STN \cite{Spatial} using $\mathrm{\Phi_{YX}}$, and direction (2): the input image Y and the output image $\mathrm{M}\textunderscore\mathrm{XY}$ which is warped from $\mathrm{A}\textunderscore\mathrm{X}$ by STN \cite{Spatial} using $\mathrm{\Phi_{XY}}$. Typically, a regularization term, such as $L_{2}$-$norm$, is applied to both $\mathrm{\Phi_{XY}}$ and $\mathrm{\Phi_{YX}}$ to control the smoothness of deformations and reduce folds, thereby guaranteeing the registration process an ideal topology-preserving transformation \cite{AnBalakrishnan}. However, the sole application of the $L_{2}$-$norm$ often fails to achieve satisfactory results. The excessive use of $L_{2}$-$norm$ as a global constraint may limit normal deformations and lead to a decrease in registration accuracy, while insufficient use cannot guarantee smoothness and bijection, making it challenging to preserve the topology. 

\subsubsection*{\textnormal{\textit{\fontsize{9}{20}\selectfont 3.2.2. The BLOC constraint}}}
To address the issue mentioned in section 3.2.1, we introduce the BLOC constraint which includes a bidirectional consistency constraint and a local anti-folding constraint. As marked with yellow dotted lines in  \textbf{\footnotesize\autoref{Fig. 2} \textcolor{blue}{(b)}}, the bidirectional consistency constraint minimizes the difference between the input image and its reversely predicted image which is warped from the output image using the inverse deformation field. Specifically, the proposed bidirectional consistency constraint minimizes the differences in the following image pairs in RegN: side (1): image $\mathrm{A}\textunderscore\mathrm{X}$ and its reversely predicted image $\mathrm{M}\textunderscore\mathrm{XY}\textunderscore\mathrm{inv}$ which is warped from the output image $\mathrm{M}\textunderscore\mathrm{XY}$ by STN using inverse deformation field $\mathrm{\Phi_{XY}^{-1}}$ of $\mathrm{\Phi_{XY}}$, and side (2): image Y and its reversely predicted image $\mathrm{M}\textunderscore\mathrm{YX}\textunderscore\mathrm{inv}$ which is warped form the output image $\mathrm{M}\textunderscore\mathrm{YX}$ by STN using the reverse deformation filed $\mathrm{\Phi_{YX}^{-1}}$ of $\mathrm{\Phi_{YX}}$. It is noteworthy that each value in the deformation field $\mathrm{\Phi_{XY}}$ and $\mathrm{\Phi_{YX}}$ respectively corresponds to the coordinate displacement of each pixel in the input images $\mathrm{A}\textunderscore\mathrm{X}$ and Y of RegN, while the values in the inverse deformation fields $\mathrm{\Phi_{XY}^{-1}}$ and $\mathrm{\Phi_{YX}^{-1}}$ should theoretically correspond to the pixel coordinate displacement of $\mathrm{M}\textunderscore\mathrm{XY}$ and $\mathrm{M}\textunderscore\mathrm{YX}$, which are obtained by warping from $\mathrm{A}\textunderscore\mathrm{X}$ and Y through the deformation fields $\mathrm{\Phi_{XY}}$ and $\mathrm{\Phi_{YX}}$. Therefore, to solve the inverse deformation fields $\mathrm{\Phi_{XY}^{-1}}$ and $\mathrm{\Phi_{YX}^{-1}}$ from $\mathrm{\Phi_{XY}}$ and $\mathrm{\Phi_{YX}}$, our study employs the deformation fields $\mathrm{\Phi_{XY}}$ and $\mathrm{\Phi_{YX}}$ to warp themselves respectively and then multiply them by -1, thereby establishing the elements' corresponding relationship between $\mathrm{\Phi_{XY}^{-1}}$ and $\mathrm{M}\textunderscore\mathrm{XY}$, as well as between $\mathrm{\Phi_{YX}^{-1}}$ and $\mathrm{M}\textunderscore\mathrm{YX}$. In doing so, the spatial transformer network (STN) is also used for the warping operation. In this design, the reversibility of the deformation field and the effectiveness of the reverse deformation field are collectively constrained to ensure the reversibility and bijection of the registration mapping, thereby achieving topology preservation. 

To further limit local folds, our study also employs the local orientation consistency constraint \cite{Fast} on the deformation fields $\mathrm{\Phi_{XY}}$ and $\mathrm{\Phi_{YX}}$ as the anti-folding constraint. For the anti-folding constraint, our method calculates the Jacobian matrix of the predicted deformation fields $\mathrm{\Phi_{XY}}$ and $\mathrm{\Phi_{YX}}$, with the computation of the Jacobian matrix referencing \textbf{\footnotesize\eqref{eq:3}}, in which $\frac{\partial \Phi_x(p)}{\partial x}$ represents the partial derivative of the x-component $\Phi_x(p)$ with respect to x at point p. Subsequently, the anti-folding constraint loss function is computed using the determinant of the Jacobian matrix. The positive Jacobian determinant leads to a bijective mapping which preserves the topology in the neighborhood while the negative Jacobian determinant results in folds. Therefore, the negative values in the Jacobian determinant of the deformation field can be used in regularizing the deformation field and then incorporated into the overall loss function of the network as a regularization term. In particular, more implementation details of the anti-folding constraint loss function are provided in section 3.4.2, under the heading 'Local anti-folding loss function'.

% \textbf{\footnotesize\eqref{eq:3}}
% \begin{tcolorbox}[colback=yellow,boxrule=0pt]
\begin{equation}
J_{\Phi} (p) = \begin{pmatrix}
\frac{\partial {\Phi_x} (p)}{\partial x} & \frac{\partial {\Phi_x} (p)}{\partial y} \\
\frac{\partial {\Phi_y} (p)}{\partial x} & \frac{\partial {\Phi_y} (p)}{\partial y}
\end{pmatrix}\label{eq:3}
\tag{3}
\end{equation}
% \end{tcolorbox}

In a word, RegN integrates the bidirectional consistency constraint and the local anti-folding constraint as an overall regularization term dubbed BLOC, which jointly helps preserve the topology of the images before and after deformation.

\subsubsection*{\textnormal{\textit{\fontsize{9}{20}\selectfont 3.2.3. Network structure}}}
As shown in \textbf{\footnotesize\autoref{Fig. 4}}, RegN is a 2D CNN architecture consisting of encoder and decoder modules with skip connections similar to VoxelMorph \cite{VoxelMorph}. The final two parallel convolutional layers generate symmetric dense displacement vector fields for bidirectional registration. RegN comprises an input layer, four downsampling blocks (DDSM), four upsampling blocks (DUSM), and two output layers. Each DUSM first establishes a skip connection between the feature maps from the previous DDSM of its same-level DDSM and the feature maps that have undergone one round of transpose convolution, followed by two convolutional layers. The last upsampling block is followed by two parallel convolutional layers serving as the network outputs. Additionally, a dual-domain attention module called Convolutional Block Attention Module (CBAM) \cite{Cbam} is incorporated at the end of the input layer and the last DUSM, thereby enhancing network performance.
\begin{figure}  % [width=\textwidth] 和 *表示两栏
	\centering
		\includegraphics[scale=.47]{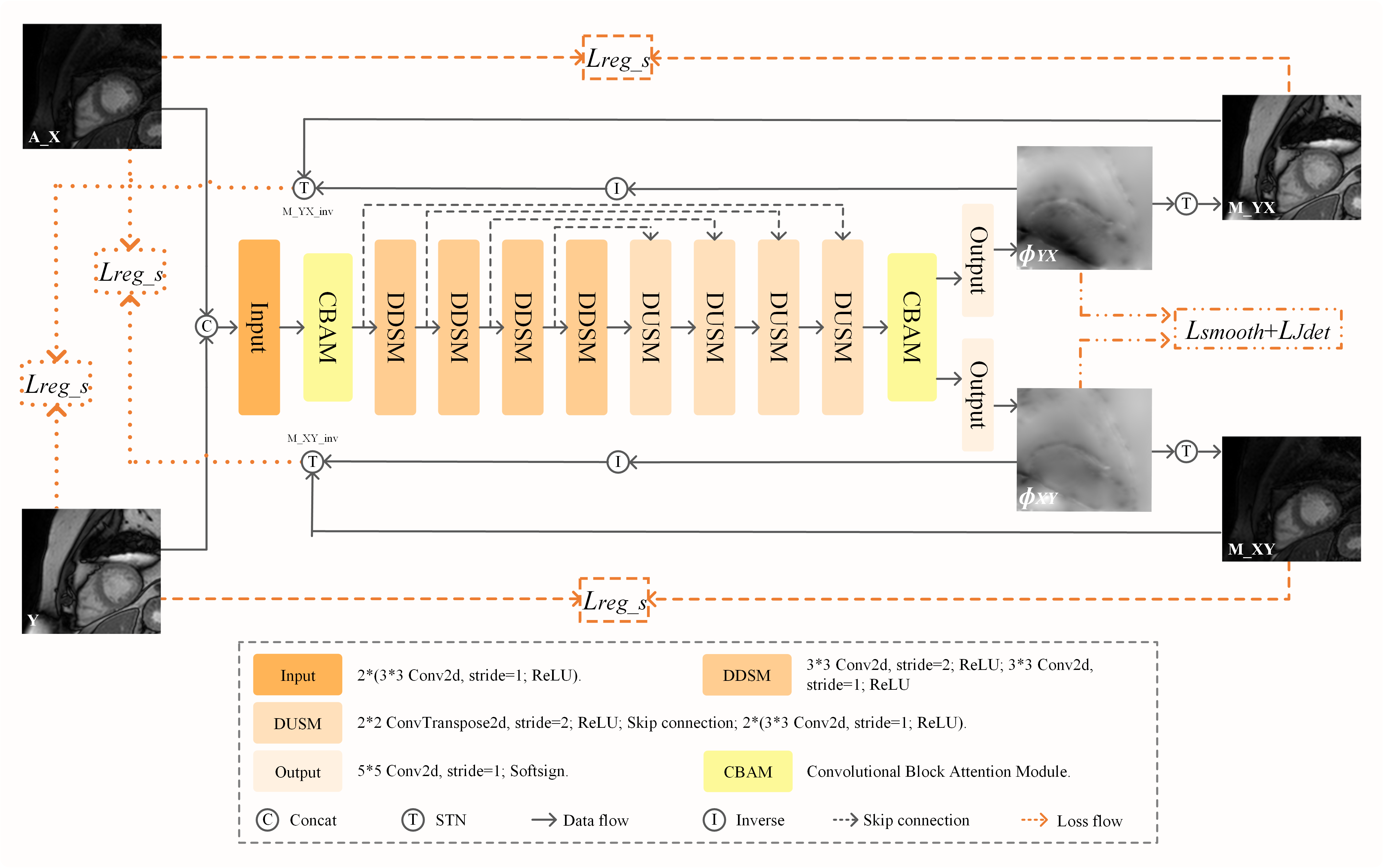}
	\caption{\textrm{Implementation details of the network architecture for RegN.}}
	\label{Fig. 4}
\end{figure}

\subsubsection*{\textnormal{\textit{\fontsize{9}{20}\selectfont 3.2.4. Weakly-supervised learning}}}
As shown in \textbf{\footnotesize\autoref{Fig. 2} \textcolor{blue}{(b-c)}}, RegN is essentially a weakly-supervised registration network, with its weak supervision information from the myocardium segmentation pseudo-labels derived from the myocardium segmentation network SegN. The specific implementation details will be elaborated in section 3.4.2.

\subsection*{\textnormal{\textit{\fontsize{10}{20}\selectfont 3.3. Myocardium segmentation network (SegN)}}}
\subsubsection*{\textnormal{\textit{\fontsize{9}{20}\selectfont 3.3.1. Network structure}}}
SegN builds upon the architecture of the widely-used U-net \cite{U-net} and utilizes a similar structure with encoder and decoder components connected through skip connections. As shown in \textbf{\footnotesize\autoref{Fig. 5}}, the encoder of SegN consists of one input layer and four downsampling blocks (SDSM), while the decoder of SegN is composed of four upsampling blocks (SUSM) and one output layer. Each SUSM connects feature maps from the previous SDSM of its same-level SDSM and feature maps which are obtained from the previous SUSM and then undergone one round of transpose convolution, followed by two convolutional layers. The final layer of SegN is a $1\times1$ convolutional layer, responsible for producing the segmentation map. Similar to previous network designs, the CBAM attention module is employed at the end of the input layer and the final SUSM.
\begin{figure}  % [width=\textwidth] 和 *表示两栏
	\centering
		\includegraphics[scale=.5]{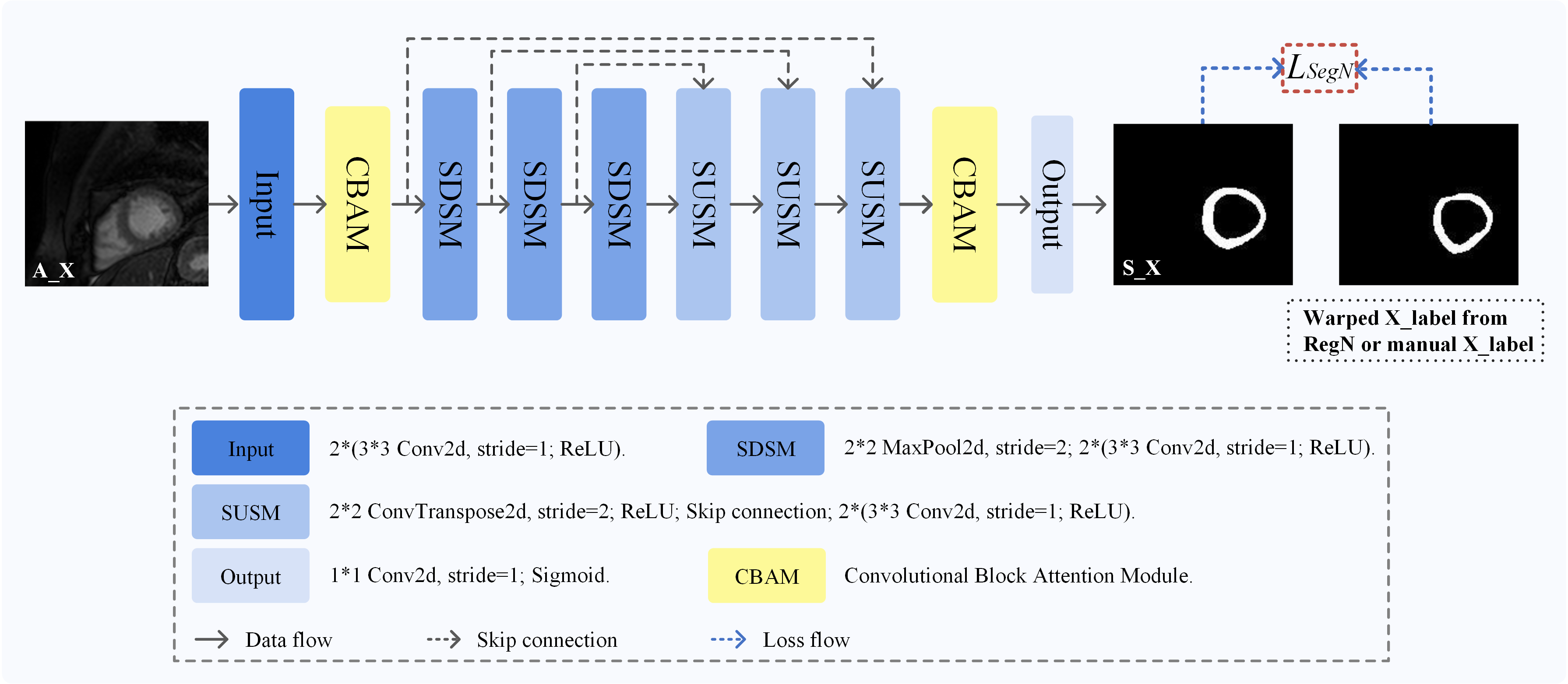}
	\caption{\textrm{Implementation details of the network architecture for SegN.}}
	\label{Fig. 5}
\end{figure}

\subsubsection*{\textnormal{\textit{\fontsize{9}{20}\selectfont 3.3.2. Semi-supervised learning}}}
SegN is a semi-supervised myocardium segmentation network whose labels are partly from manual myocardium segmentation labels and partly from deformed manual myocardium segmentation labels from RegN, with the loss function defined in section 3.4.3.
\end{small}

\subsection*{\textnormal{\textit{\fontsize{10}{20}\selectfont 3.4. Loss function.}}}
\subsubsection*{\textnormal{\textit{\fontsize{9}{20}\selectfont 3.4.1. Loss function of AffN}}}
\begin{small}
Mean-Square Error ($\mathit{\textit{MSE}}$), Normalized Cross Correlation($\mathit{\textit{NCC}}$), Mutual Information ($\mathit{\textit{MI}}$) \cite{MI}, Normalized Gradient Fields ($\mathit{\textit{NGF}}$) \cite{Cardiac, Intensity}, and Modality Independent Neighborhood Descriptor ($\mathit{\textit{MIND}}$) \cite{MIND}, are typically used as similarity loss functions in deep learning-based image registration methods. In our study, we treated the registration of cardiac T1w images as a multimodal registration and proposed a weighted image similarity metric, $\mathit{\textit{WLs}}$, to address significant contrast variation issue by simultaneously considering signal intensity and structural variations in the registration pair. Considering the contrast variations in cardiac T1w images, $\mathit{\textit{NCC}}$ and $\mathit{\textit{MI}}$ are selected to represent intensity similarity, while $\mathit{\textit{NGF}}$ and $\mathit{\textit{MIND}}$ are chosen to represent structural similarity.

Specifically, the expression for $\mathit{\textit{WLs}}$ is defined as \textbf{\footnotesize\eqref{Eq:4}}. Weight $a$, $b$, $c$, and $d$ represent the weights of $\mathit{\textit{NCC}}$, $\mathit{\textit{MI}}$, $\mathit{\textit{NGF}}$, and $\mathit{\textit{MIND}}$ respectively. $I$ and $J$ are the registration pair.

\begin{equation}
\begin{split}
\mathit{\textit{WLs}(I, J)} &= a \cdot NCC(I, J)+b \cdot MI(I, J) \\
&\quad + c \cdot NGF(I, J)+d \cdot MIND(I, J)\label{Eq:4}
\end{split}
\tag{4}
\end{equation}
\end{small}

\begin{small}
Consequently, the overall loss function of AffN is:
\begin{equation}
\mathit{L_{\text{\textit{AffN}}}} = - \textit{WLs}(A\textunderscore{X},Y)
\tag{5}
\end{equation}

\subsubsection*{\textnormal{\textit{\fontsize{9}{20}\selectfont 3.4.2. Loss function of RegN}}}
\textit{Image similarity loss function:}  Image similarity loss function encompasses two components: the baseline design and the BLOC constraint. $- \textit{WLs}(\textit{A}\textunderscore{\textit{X}},\textit{M}\textunderscore{\textit{YX}}) - \textit{WLs}(Y,\textit{M}\textunderscore{\textit{XY}})$ is associated with the baseline design. Meanwhile, $- \textit{WLs}(Y,\textit{M}\textunderscore{\textit{YX}}\textunderscore{\textit{inv}}) - \textit{WLs}(\textit{A}\textunderscore{\textit{X}},$ $\textit{M}\textunderscore{\textit{XY}}\textunderscore{\textit{inv}})$ corresponds to the bidirectional consistency constraint within the BLOC constraint. Similar to AffN, WLs is also used as the image similarity metric in RegN. As described in section 3.2, the image similarity loss function of RegN is expressed as follows:

\begin{equation}
\begin{split}
\textit{Lreg\textunderscore{s}} &= - \textit{WLs}(Y,\textit{M}\textunderscore{\textit{YX}}\textunderscore{\textit{inv}}) \\
&\quad - \textit{WLs}(\textit{A}\textunderscore{\textit{X}},\textit{M}\textunderscore{\textit{XY}}\textunderscore{\textit{inv}}) \\
&\quad - \textit{WLs}(\textit{A}\textunderscore{\textit{X}},\textit{M}\textunderscore{\textit{YX}}) - \textit{WLs}(Y,\textit{M}\textunderscore{\textit{XY}}) \\
\end{split}
\tag{6}
\end{equation}
\end{small}

\begin{small}
\textit{Weakly-supervised loss function:}  We utilize the myocardium segmentation results obtained from SegN as weak supervision information for RegN. The soft multi-class dice coefficient ($\mathit{\textit{D}}$) \cite{DeepAtlas} expressed at \textbf{\footnotesize\eqref{Eq:7}} is used to measure the similarity of myocardium segmentation. It should be noted that the pseudo-labels used in RegN are all sourced from the myocardium segmentation outcomes of SegN rather than manually segmented labels. 
% \begin{equation}
% \hat{A_{\theta}} = \text{argmin}\mathcal{L}_{sim}(f, \mathcal{F}_{A}(m, A_{\theta}))
% \tag{2}
% \end{equation}

\begin{equation}
\mathcal{D}(S_{I}, S_{J}) = 1 - \frac{1}{k}\sum_{k=1}^{K} \frac{{\sum_{}}^{}_{x}{S_{I}}_{k}(x){{S}_{J}}_{k}(x)}    {{\sum_{}}^{}_{x}{{S}_{I}}_{k}(x)+{\sum_{}}^{}_{x}{{S}_{J}}_{k}(x)} \label{Eq:7}
\tag{7}
\end{equation}

\begin{equation}
\begin{split}
\mathit{\textit{{L}{reg\textunderscore{a}}}} &= \mathcal{D}_{r}(S\_{X}\circ \mathrm{\Phi_{XY}},S\_Y) \\
&\quad +\mathcal{D}_{r}(S\_{Y}\circ \mathrm{\Phi_{YX}},S\_{X})\label{Eq:8}
\end{split}
\tag{8}
\end{equation}

The weakly-supervised label similarity loss function, $Lreg\textunderscore{a}$, is designed for RegN in \textbf{\footnotesize\eqref{Eq:8}}, where $\mathcal{D}_{r}(\cdot, \cdot)$ computes the dice loss between the myocardium segmentation outcome (segmented by SegN) of image Y and the deformed myocardium segmentation outcome (segmented by SegN and then warped by RegN) of image $\mathrm{A}\textunderscore{\mathrm{X}}$, the other direction likewise. $\mathrm{S}\textunderscore{\mathrm{X}}$ and $\mathrm{S}\textunderscore{\mathrm{Y}}$ are the myocardium segmentation outcomes of $\mathrm{A}\textunderscore{\mathrm{X}}$ and Y in SegN, $\circ$ represents the warping performed through STN \cite{Spatial} and $\mathrm{\Phi_{XY}}$ represents the registration mapping predicted by RegN.

\textit{Local anti-folding loss function:}
To address the issue of local folding, local orientation consistency constraint $\mathit{L_{\text{\textit{Jdet}}}}$ \cite{Fast} is also used with the expression as follows:

\begin{equation}
\mathit{L_{\text{\textit{Jdet}}}} = \frac{1}{N}\sum_{p \in \Omega} \sigma(-|J_{\Phi_{XY}}(p)|)+\frac{1}{N}\sum_{p \in \Omega} \sigma(-|J_{\Phi_{YX}}(p)|) \\
\tag{9}
\end{equation}

Where $|J_{\Phi_{XY}}(p)|$ represents the determinant of the Jacobian matrix of deformation field $\Phi_{XY}$ at position \textit{p}. Therefore, an activation function, $\sigma(\cdot)$, which maintains the positive value and makes the negative value zero, is utilized to penalize the regions that are not topology-preserving. $N$ denotes the total number of elements in $|J_{\Phi_{XY}}(p)|$. The local anti-folding constraint allows for a more adaptive focus on preserving the topology at a local level rather than considering the global smoothness. Therefore, the degree of topology preservation can be improved by combining it with the proposed bidirectional consistency constraint to be an overall constraint, BLOC. By working with $L_{2}$-$norm$ which is marked as $\mathit{L_{\text{\textit{smooth}}}}$, a balance can be achieved between registration accuracy and deformation field smoothness.

The overall loss function of RegN is defined in \textbf{\footnotesize\eqref{Eq:10}}, $\lambda_{1}$ and $\lambda_{2}$ are the weights of the anti-folding constraint and the $L_{2}$-$norm$.
\begin{equation}
\mathit{L_{\text{\textit{RegN}}}} = \textit{Lreg\textunderscore{s}} + \textit{Lreg\textunderscore{a}}+\lambda_{1}\mathit{L_{\text{\textit{Jdet}}}} + \lambda_{2}\mathit{L_{\text{\textit{smooth}}}}\label{Eq:10}
\tag{10}
\end{equation}

\subsubsection*{\textnormal{\textit{\fontsize{9}{20}\selectfont 3.4.3. Loss function of SegN}}}
\textit{Semi-supervised loss function:}
The label similarity metric expressed at \textbf{\footnotesize\eqref{Eq:7}} is also used in the overall training loss function ($L_{SegN}$) of SegN.

%\begin{normalsize}
\begin{equation}
\scriptstyle{L_{SegN}} = \left\{\vphantom{\begin{array}{ll} y_2 \\ y_3 \end{array}} \right.\!
\begin{array}{ll}
    \scriptstyle{\lambda_{r}\mathcal{D}_{r}(X_{l}\circ\mathrm{\Phi_{XY}}, S\textunderscore{Y}) +  \lambda_{s}\mathcal{D}_{s}(S\textunderscore{X},X_{l})},\,\scriptstyle{\text{if}\, A\textunderscore{X}\, is\, labeled}. \\
    \scriptstyle{\lambda_{r}\mathcal{D}_{r}(Y_{l}\circ\mathrm{\Phi_{YX}}, S\textunderscore{X}) +  \lambda_{s}\mathcal{D}_{s}(S\textunderscore{Y},Y_{l})},\,\scriptstyle{\text{if}\, {Y}\, is\, labeled}. \\
    \scriptstyle{\lambda_{r}\mathcal{D}_{r}(Y_{l}\circ\mathrm{\Phi_{YX}}, S\_X) +  \lambda_{s}\mathcal{D}_{s}(S\textunderscore{Y},Y_{l})},\,\scriptstyle{\text{if} \,A\textunderscore{X}\, and\, {Y}\, are\, labeled}.
\end{array}\label{11}
\tag{11}
\end{equation}
%\end{normalsize}
\end{small}

\begin{small}
Where $\mathrm{X}_\mathrm{l}$ and $\mathrm{Y}_\mathrm{l}$ represent manual myocardial segmentation labels for $\mathrm{A\_X}$ and $\mathrm{Y}$. $\mathcal{D}_{r}(\cdot, \cdot)$ represents the dice coefficient between the deformed myocardium segmentation outcome (segmented manually and then warped by RegN) and the myocardium segmentation outcome (segmented by SegN), while $\mathcal{D}_{s}(\cdot, \cdot)$ represents the dice coefficient between the myocardium segmentation outcome (segmented by SegN) and the manual myocardium segmentation label. The combination of $\mathcal{D}_{r}(\cdot, \cdot)$ part and $\mathcal{D}_{s}(\cdot, \cdot)$ part makes SegN a semi-supervised network. $\lambda_{r}$ and $\lambda_{s}$ are the weights of unsupervised and supervised learning.    
\end{small}

\subsection*{\textnormal{\textit{\fontsize{10}{20}\selectfont 3.5. Network training}}}
\begin{small}
The training of RS-MOCO can be summarized in the following steps:
\begin{enumerate}
\itemsep=0pt
\item  AffN is firstly employed for pre-training to achieve initial linear affine registration and correct partial misalignment caused by motion. Then, the pre-trained AffN model is integrated into the deformable registration network as its pre-alignment module;
\item  RegN is utilized for deformable registration training to correct the vast majority of misalignment and achieve more precise registration results;
\item  SegN is used for myocardium segmentation training to obtain pseudo-labels of myocardium segmentation. It is worth noting that SegN is semi-supervised, with some pseudo-labels derived from the distorted ground truth labels through the trained RegN;
\item  Re-train the weakly-supervised RegN using the pseudo-labels of myocardium segmentation generated by SegN;
\item  Repeating steps 3 and 4, RegN and SegN are trained alternatively, to continuously improve the performance of both networks.
\end{enumerate}
\end{small}

\section{Results}
\begin{small}
\subsection*{\textnormal{\textit{\fontsize{10}{20}\selectfont 4.1. Datasets}}}
This study utilized the publicly available dataset, T1Dataset210 \cite{Nonrigid}, from Harvard Dataverse. Imaging was performed using a 1.5T Philips Achieva system (Philips Healthcare, Best, Netherlands) with a 32-channel cardiac coil. T1 mapping was performed in 210 consecutive patients (134 males; age 57±14 years) with known or suspected cardiovascular diseases referred for a clinical cardiac MR exam. The imaging protocol was the free-breathing, respiratory-navigated, slice-interleaved T1 mapping sequence \cite{STONE}. Each patient data consisted of five short axial slices covering the LV from base to apex. At each slice location, 11 T1w images were acquired at different inversion times, TI$_{i}$. The epicardial and endocardial boundaries of all images in the database (N = 11550 images) were manually delineated, with each contour starting from a myocardium point closest to the anterior insertion of the right ventricle into the LV.

Furthermore, the generalization performance of different methods was tested on an external dataset, SZC-T1, which was obtained at Shenzhen Children's Hospital. The SZC-T1 dataset consisted of cardiac T1 mapping images from 8 patients(6 males; age 10±3 years) acquired using the MOLLI sequence \cite{Modified} on a 3.0T SIEMENS Skyra with an 18-channel cardiac coil. Patients underwent respiratory training before scanning, and all images were acquired under free-breathing using a prospective electrocardiogram gating technique. Each patient data consisted of 8-9 short axial slices covering the LV from base to apex with each slice comprising of 8 T1w images acquired at different inversion times. 
    
\subsection*{\textnormal{\textit{\fontsize{10}{20}\selectfont 4.2. Implementation}}}
\subsubsection*{\textnormal{\textit{\fontsize{9}{20}\selectfont 4.2.1. Execution specifics}}}
During training and testing, we crop each T1w image in the internal dataset from $320$*$320$ to a size of $144$*$160$ and perform normalization. All models are trained using the Adam optimizer with an initial learning rate of 1e-4 for affine registration, 1e-3 for deformable registration and segmentation, and a batch size of 16. The parameter sizes of the three networks, AffN, RegN, and SegN, are approximately 1020k, 1517k, and 1946k respectively. During the training of one batch of data, the GPU memory usage of these networks is approximately 4238MiB, 5104MiB, and 3306MiB respectively. Considering both model performance and computational cost, training was halted when no further improvement in performance on the validation set was observed in order to prevent overfitting and simultaneously eliminate unnecessary computational expenses. Based on our experimental experience, AffN achieves optimal performance around five epochs of training. Furthermore, the overall model reaches peak performance when RegN and SegN are alternately trained at a ratio of 2 epochs to 0.5 epochs and this process is repeated three times. The weights of different components in $\mathit{\textit{WLs}}$ were determined through experiments using the hyperparameter optimization tool, Optuna. Based on the experimental results, the weights for $\mathit{\textit{WLs}}$ in this study were set as: 1.1, 4, 3.3 and 8.3, respectively. Also, the weights for the Jacobian determinant regularization loss and the $L_{2}$-$norm$ loss of the deformation field are set to 1000 and 8 according to Optuna. We conduct experiments using triple fold cross validation, with each experiment having 92,400 registration pairs available for training. All experiments are based on PyTorch 1.11.0 and deployed on a server equipped with an RTX A6000 GPU.

\subsubsection*{\textnormal{\textit{\fontsize{9}{20}\selectfont 4.2.2. Evaluation metrics}}}

This study quantitatively evaluates the motion correction effect of cardiac T1 mapping using the Dice Similarity Coefficient ($\mathit{\textit{DSC}}$) which quantifies the degree of overlap between the predicted myocardial segmentation labels and the manual myocardial segmentation labels, Hausdorff Distance ($\mathit{\textit{HD}}$) which quantifies the similarity of myocardial contours between the predicted and manual myocardial segmentation labels, and the number of non-positive Jacobian determinants ($|J_{\Phi}|$$\leq$$0$) which quantifies the number of pixels within a deformation field that is in a state of folding. The smaller value of $|J_{\Phi}|$$\leq$$0$ indicates better topology preservation performance. For qualitative evaluation, a three-parameter signal model is employed to fit the T1 map for each STONE or MOLLI series. During testing, the 11th or 8th T1w image of each STONE or MOLLI series is selected as the fixed image, while the remaining images serve as moving images. Additionally, the testing time is measured for different methods to compare the efficiency of motion correction in each STONE or MOLLI series.

\subsubsection*{\textnormal{\textit{\fontsize{9}{20}\selectfont 4.2.3. Comparison methods}}}
We employed the Cb-Reg scheme (a part of work by El‐Rewaidy et al. \cite{Nonrigid}), which utilizes manually delineated myocardial contours for local registration, to register the internal dataset T1Dataset210 and external dataset SZC-T1. It is noteworthy that Cb-Reg executes registration utilizing manually delineated myocardial contours, rather than using myocardial contours automatically extracted by the active shape model. In addition, this study compares various medical image registration and motion correction methods for cardiac T1 mapping. We implemented the classic symmetric image normalization method (SyN) \cite{TheAvants, AAvants} using the Python-based ANTs package. We compared our approach with DisQ (DisQ) \cite{DisQ}, a method specifically designed for motion correction in cardiac T1 mapping, proposed by Yang et al. Additionally, we utilized VoxelMorph (VM) \cite{VoxelMorph}, a widely used unsupervised registration method based on deep learning, for intensity-based registration.

\subsubsection*{\textnormal{\textit{\fontsize{9}{20}\selectfont 4.2.3. Ablation study}}}
Our research involves extensive ablative experiments. The first part aims to validate the effectiveness of $\mathit{\textit{WLs}}$ by comparing its results with those obtained by $\mathit{\textit{NCC}}$, $\mathit{\textit{MI}}$, $\mathit{\textit{NGF}}$, and $\mathit{\textit{MIND}}$ individually used in image similarity loss function of AffN and RegN. The second part aims to verify the effectiveness of each module in RS-MOCO. Specifically, we test and compare the network results with each module removed separately, including AffN, SegN, CBAM, and BLOC.

\subsection*{\textnormal{\textit{\fontsize{10}{20}\selectfont 4.3. Motion correction of cardiac T1 mapping}}}
\subsubsection*{\textnormal{\textit{\fontsize{9}{20}\selectfont 4.3.1. Qualitative results analysis}}}
\textit{Motion correction effects on T1w images:} \textbf{\footnotesize\autoref{Fig. 6}} presents a comparative analysis of the effectiveness of different methods for motion correction on the internal dataset, T1dataset210 \cite{Nonrigid}. Prior to motion correction, each T1w image exhibits noticeable drift motion compared to the Ref image, as indicated by the white arrows. After motion correction, the degree of contour overlap between each T1w image and the "Ref" T1w image demonstrates varying levels of improvement. Specifically, SyN, DisQ, and VoxelMorph can only correct partial frames and are ineffective in addressing larger displacements. While our method and Cb-Reg achieve the best results, with the majority of misalignment in the T1w images corrected. Furthermore, as shown in \textbf{\footnotesize\autoref{Fig. 6}\textcolor{blue}{(1-3)}}, frames with significantly reduced signal intensity and larger displacements are also effectively corrected. These results demonstrate the capability of our method in addressing the issue of significant contrast variations in cardiac T1 mapping.
\begin{figure*} % [width=\textwidth] 和 *表示两栏
	\centering
	   \includegraphics[width=0.68\textwidth]{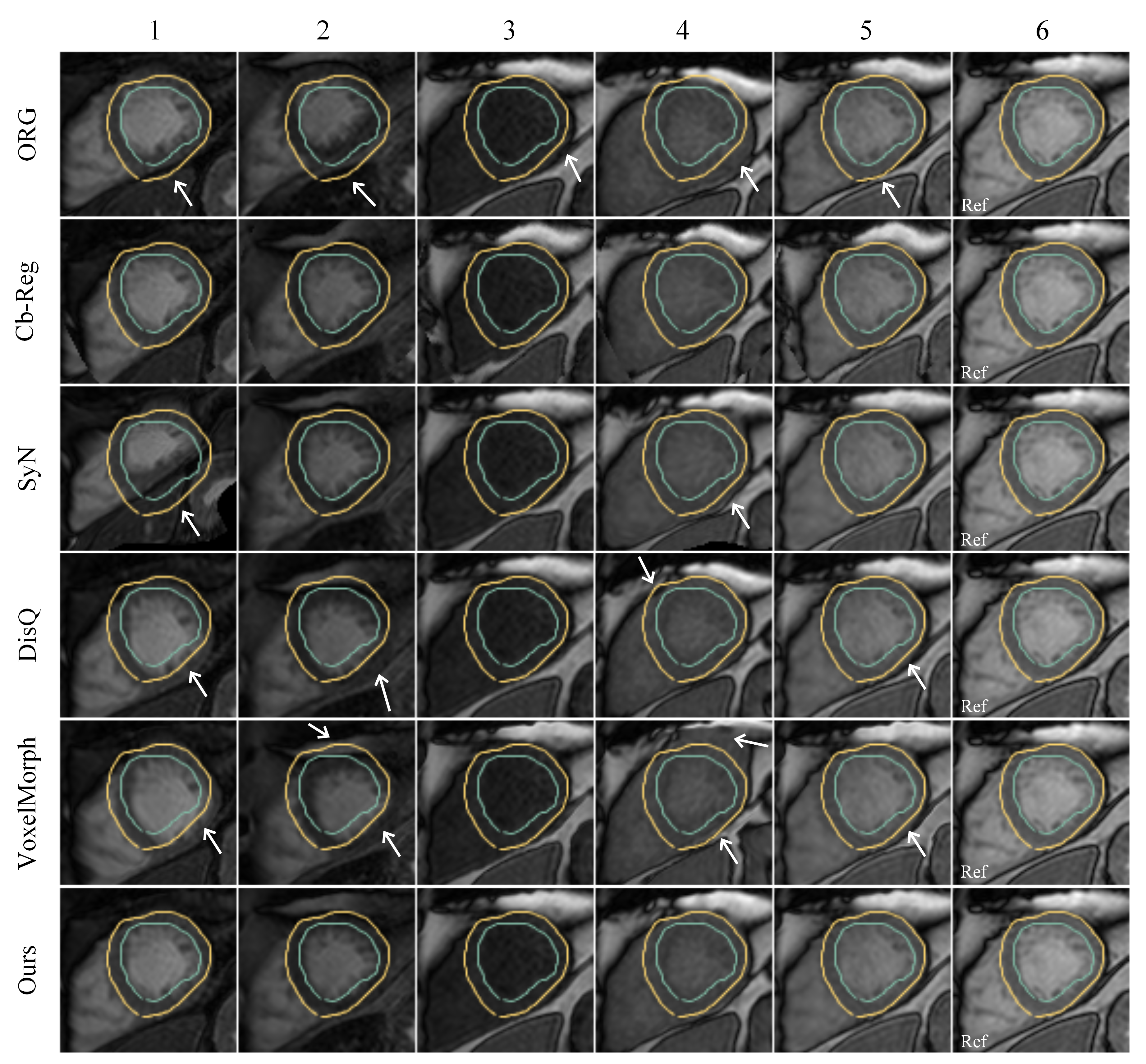}
    \caption{\textrm{Comparison of T1w images before and after correction on T1dataset210. For demonstration purposes, six images from a specific STONE series were selected. Among them, ORG represents the uncorrected T1w images, while the rest are T1w images after correction using different methods. The mark, Ref, denotes the reference frame used for registration. Additionally, the manual epicardial and endocardial contours of the reference frame were overlaid on the remaining T1w images to illustrate the registration effect.}}
    \label{Fig. 6} 
\end{figure*}\
% To facilitate visual inspection of registration results, we applied different display ranges to different T1w images from the same series.

\textbf{\footnotesize\autoref{Fig. 7}} presents the performance of different methods on the external dataset, SZC-T1. Due to the more consistent nature of children's respiration, most original ROI misalignments were smaller compared to the T1dataset210 dataset, making the registration task relatively simpler. However, the contrast of SZC-T1, acquired using a 3.0T magnetic resonance scanner, experienced significant changes compared to T1dataset210 acquired by a 1.5T magnetic resonance scanner. Specifically, within the SZC-T1 dataset, the contrast difference between the myocardium signal and the surrounding tissue signals (such as the blood pool and fat) is substantially reduced, thereby increasing the difficulty in registration and network generalization. The ORG row in \textbf{\footnotesize\autoref{Fig. 7}} illustrates the varying degrees of ROI misalignment in position and shape present in the unregistered T1w images. Even for small misalignments, VoxelMorph demonstrated limited correction capability. SyN and DisQ yield slightly better results than VoxelMorph, but they still fail to eliminate misalignments completely. Additionally, while Cb-Reg exhibits good performance on internal dataset, its performance decreases on external dataset. This is due to Cb-Reg’s reliance entirely on endocardial and epicardial contour information for registration, whereas the incorrectly delineated contours caused by poorer image quality (such as indistinctness of the endocardial and epicardial boundaries and low image resolution) can result in registration failure. This reflects its limited flexibility and adaptability across varying datasets. In contrast, our method exhibits better adaptability on the external dataset, correcting the majority of misalignment. Notably, the third frame with the largest tissue contrast variation achieved good registration results. Moreover, our method outperforms SyN in the correction of myocardial boundaries, demonstrating the correction and generalization capability of our approach.

\begin{figure*} % [width=\textwidth] 和 *表示两栏
	\centering
	   \includegraphics[width=0.68\textwidth]{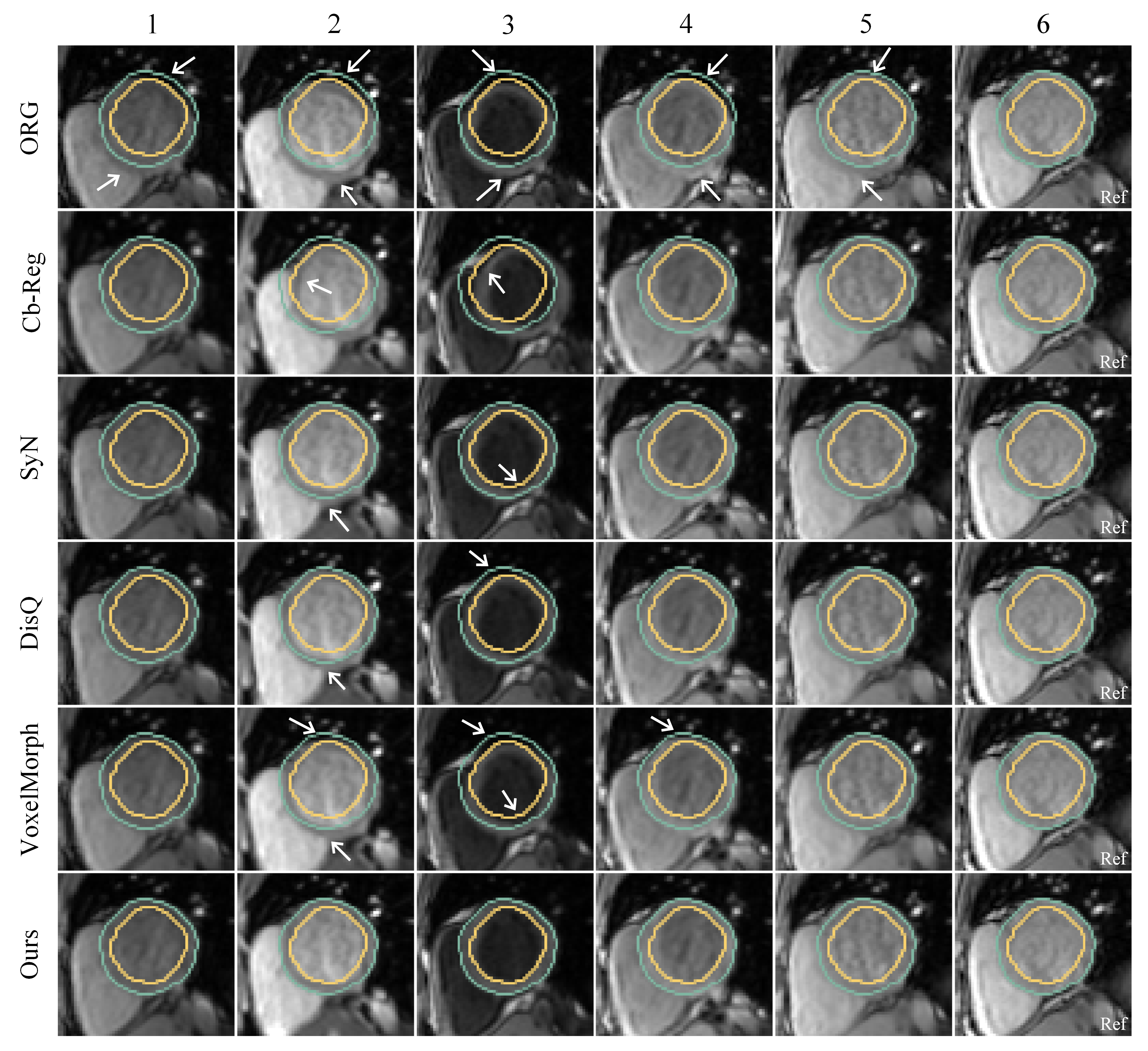}
    \caption{\textrm{Comparison of T1w images before and after correction on SZC-T1. For demonstration purposes, six images from a specific MOLLI series were selected. The epicardial and endocardial contours of the "Ref" image are synchronized onto the remaining T1w images to demonstrate the correction effect.}}
    \label{Fig. 7} 
\end{figure*}

\textit{Motion correction effects on T1 maps:} 
Three examples of the fitted T1 maps before and after motion correction on the internal dataset, T1dataset210, using different methods are shown in \textbf{\footnotesize\autoref{Fig. 8}}. "ORG" represents the T1 maps without motion correction, exhibiting significant motion artifacts leading to severe signal confusion in the myocardial region. With the application of our method, RS-MOCO, for motion correction, the quality of the T1 maps is significantly improved, and the myocardial region becomes more homogeneous. These three examples demonstrate that RS-MOCO can correct motion artifacts of different degrees and exhibit good stability. The correction effect of Cb-Reg is similar to our method. Furthermore, the conventional SyN method is only capable of accurately correcting frames within the STONE series that have similar contrast to the reference image. However, it is ineffective when dealing with frames that exhibit significant contrast differences. Additionally, SyN demonstrates poor robustness and frequently results in significant registration errors. The DisQ method carries the risk of errors in both the disentangled representation learning and image registration steps, consequently hindering its ability to achieve robust motion correction for all frames in a STONE series, especially in the presence of severe motion artifacts. The VoxelMorph framework, based on deep learning, fails to correct frames with significant contrast variations compared to the reference image due to the use of a single image similarity metric in registration. 

\begin{figure*} % [width=\textwidth] 和 *表示两栏
	\centering
	   \includegraphics[width=0.8\textwidth]{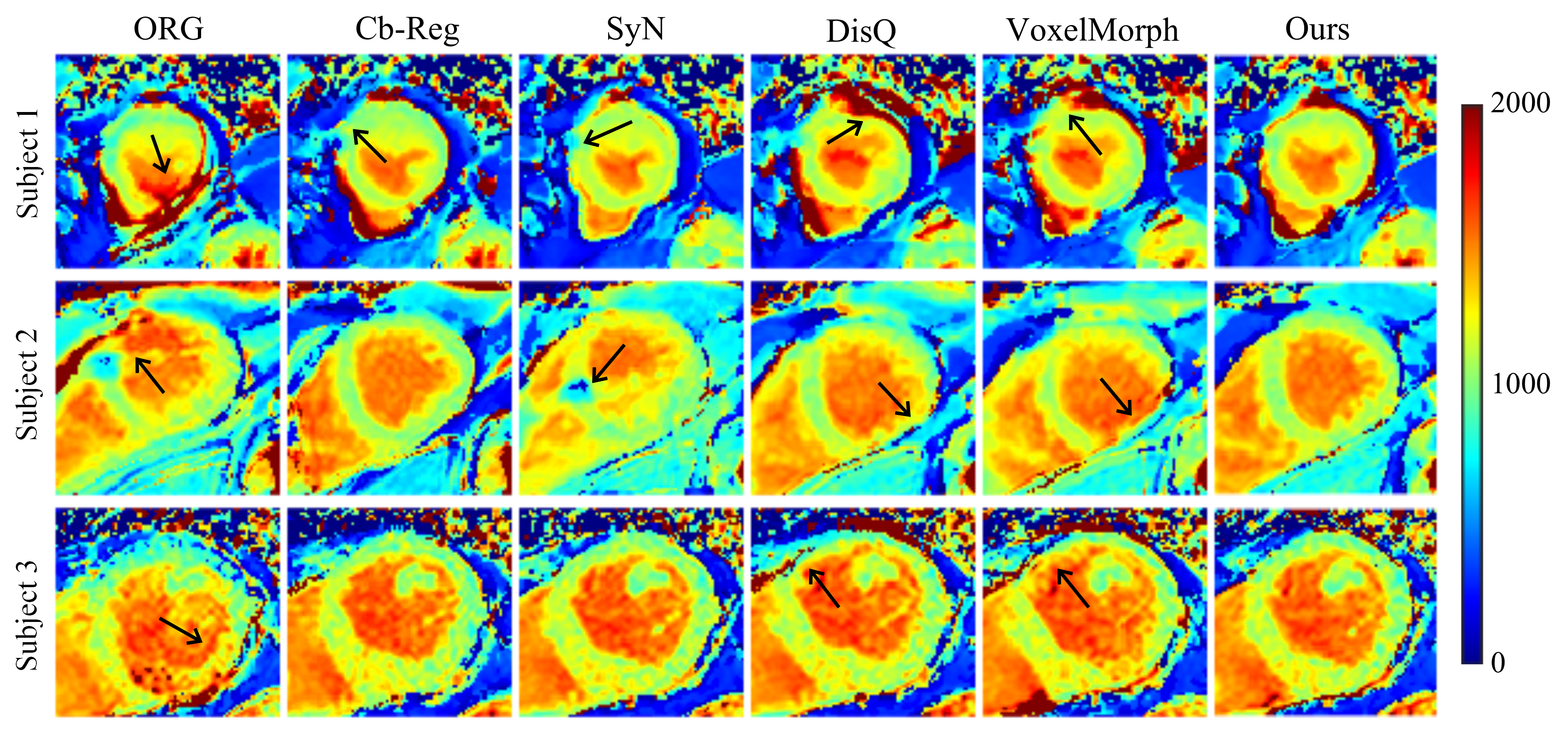}
    \caption{\textrm{Comparison of uncorrected T1 maps (ORG) and corrected T1 maps from T1dataset210 using various methods: Cb-Reg, SyN, DisQ, VoxelMorph, and our proposed method RS-MOCO.}}
    % \textrm{\footnotesize{\textbf{Fig. 2.} The overall workflow diagram of RS-MOCO.}}
    \label{Fig. 8} 
\end{figure*}

\textbf{\footnotesize\autoref{Fig. 9}} illustrates three example subjects from the external dataset, SZC-T1. From the ORG row which denotes the T1 maps without motion correction, it can be observed that motion artifacts in SZC-T1 are significantly reduced compared to the T1dataset210 dataset. However, VoxelMorph still exhibits poor correction and generalization performance, as it is unable to effectively correct motion artifacts and even introduces new artifacts. DisQ demonstrates better correction ability compared to VoxelMorph. However, some residual uncorrected areas are observed in certain locations of the endocardium and epicardium, which may be attributed to its limited capability in contrast decomposition, thereby being unable to effectively extract the faint signals in the region of endocardium and epicardium. SyN exhibits improved robustness on the external SZC-T1 dataset compared to the internal T1dataset210 dataset, which may be attributed to the reduction of severe misalignments. However, it performs poorly at the myocardial edges, which may be attributed to the minimal contrast differences between certain myocardial regions and the surrounding tissues in the SZC-T1 dataset. Furthermore, the inferior motion correction performance of Cb-Reg on external dataset is associated with the observed blurring of boundaries between the myocardium and adjacent tissues in these images. In contrast, our method RS-MOCO outperforms all other methods in terms of both correction performance and generalization ability, with a more uniform signal in the myocardial region. This improvement may be attributed to the inclusion of a semi-supervised myocardium segmentation module in the network, where the added myocardium segmentation labels better guide the network in aligning the myocardial region.
\begin{figure*} % [width=\textwidth] 和 *表示两栏
	\centering
	   \includegraphics[width=0.78\textwidth]{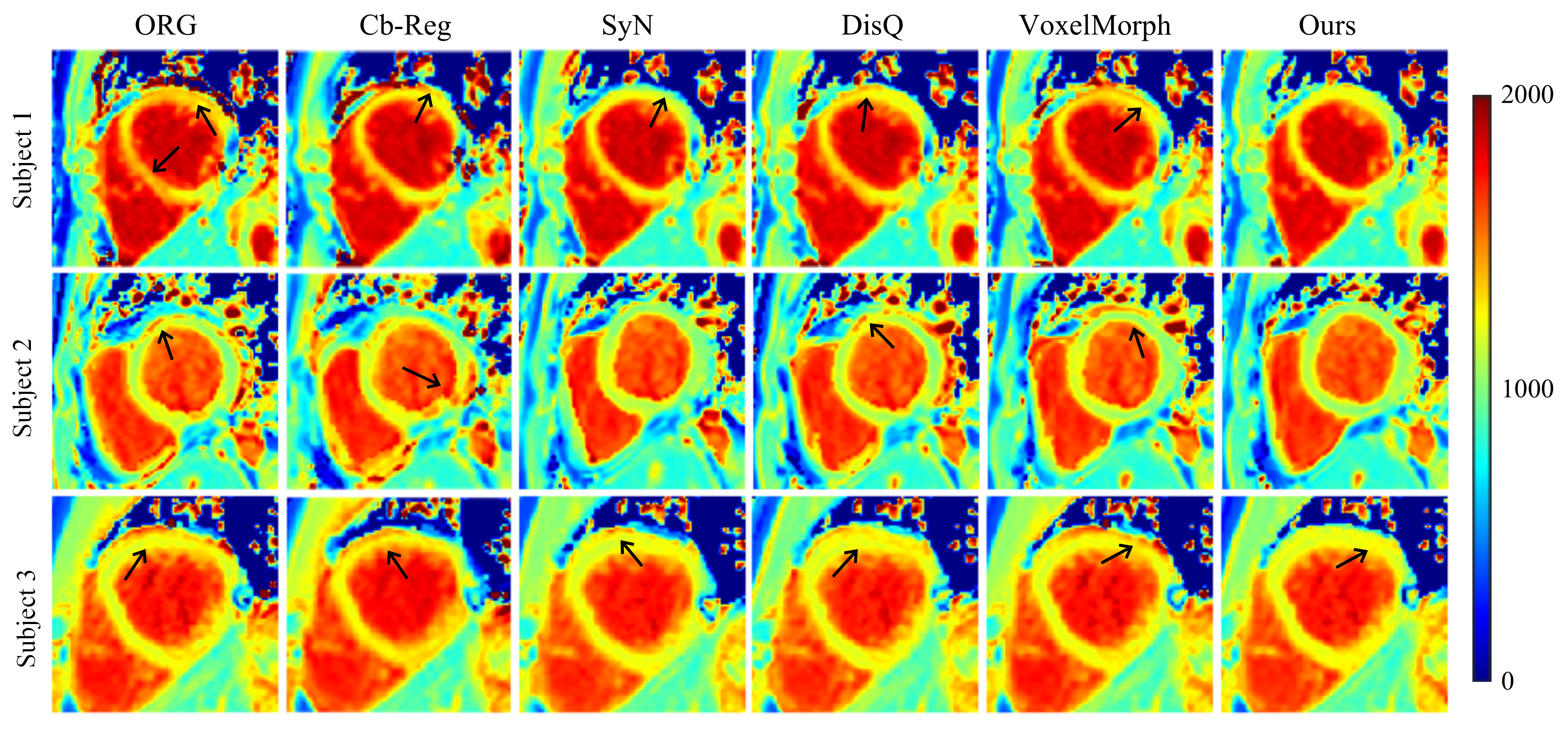}
    \caption{\textrm{Comparison of uncorrected T1 maps (ORG) and corrected T1 maps from SZC-T1 dataset using various methods: Cb-Reg, SyN, DisQ, VoxelMorph, and our proposed method RS-MOCO.}}
    % \textrm{\footnotesize{\textbf{Fig. 2.} The overall workflow diagram of RS-MOCO.}}
    \label{Fig. 9} 
\end{figure*}

\subsubsection*{\textnormal{\textit{\fontsize{9}{20}\selectfont 4.3.2. Quantitative results analysis}}}
\textbf{\footnotesize\autoref{tbl1}} presents a comparison of various evaluation metrics results on the internal dataset, T1Dataset210, before and after calibration using different methods. Among all global registration methods, our proposed method RS-MOCO achieved the optimal results with a Dice Similarity Coefficient ($\mathit{\textit{DSC}}$) of $0.827$, an endocardial Hausdorff Distance ($\mathit{\textit{HD\_endo}}$) of $6.19$ and an epicardial Hausdorff Distance ($\mathit{\textit{HD\_epi}}$) of $6.71$. In comparison to the conventional method (SyN), RS-MOCO demonstrates a significant improvement in performance and robustness, with the $\mathit{\textit{DSC}}$ improved by $7.5\%$, the endocardial $\mathit{\textit{HD\_endo}}$ decreased by $14.5\%$ and the epicardial $\mathit{\textit{HD\_epi}}$ decreased by $11.5\%$. From the perspective of topology preservation performance, with the BLOC constraint, the deformation field predicted by RS-MOCO contains only a very small number ($|J_{\Phi}|$$\leq$$0$: 0.921) of occurrences of folding. RS-MOCO achieves performance close to the traditional topology-preserving registration method SyN, while it significantly outperforms the non-topology-preserving methods DisQ and VM in terms of decreasing the number of pixels with non-positive Jacobian values by 99.56\% and 99.51\% respectively. However, compared with the contour-based local registration method Cb-Reg, the registration accuracy of our proposed RS-MOCO still needs to be improved. But since the deep learning-based RS-MOCO does not rely on the manually delineated myocardial labels at the testing stage and the processing speed is greatly improved for about 20 times, we believe that RS-MOCO still has a relatively good application prospect in the future.

\begin{table*}[width=1.25\linewidth,cols=6,pos=h]\scriptsize
\captionsetup[table]{labelfont=rm,textfont={rm,rm}}
\rm\caption{\rm Performance comparisons of different registration methods on the T1dataset210 dataset. Standard deviations are shown within parentheses and are applied to the subsequent tables as well.}\label{tbl1}
\renewcommand\arraystretch{1.5}
\begin{tabular*}{\tblwidth}{@{} cccccc@{} }
\toprule
\textrm{Method} & \textrm{DSC} & $|J_{\Phi}|$$\leq$$0$ & \textrm{HD\_endo(px)} & \textrm{HD\_epi(px)} & \textrm{Time(s)} \\
\midrule
\textrm{ORG} & $0.680(0.194)$ & $-$ & $7.80(4.02)$ & $8.03(3.79)$ & $-$ \\
\textrm{Cb-Reg} & $0.907(0.019)$ & $-$ & $5.78(2.27)$ & $6.41(2.11)$ & $224.3(9.0)$ \\
\textrm{SyN} & $0.769(0.155)$ & $0.324(0.365)$ & $7.24(3.51)$ & $7.58(3.32)$ & $376.8(16.1)$ \\
\textrm{DisQ} & $0.810(0.129)$ & $209.18(195.4)$ & $6.36(3.38)$ & $6.84(3.16)$ & $8.5(0.9)$ \\
\textrm{VM} & $0.785(0.128)$ & $187.05(164.6)$ & $6.62(2.77)$ & $7.00(2.57)$ & $9.0(0.7)$ \\
\textrm{Ours(AffN)} & $0.773(0.121)$ & $-$ & $6.74(2.59)$ & $7.13(2.52)$ & $3.3(0.2)$ \\
\textrm{Ours} & $0.827(0.093)$ & $0.921(0.552)$ & $6.19(2.06)$ & $6.71(1.95)$ & $8.1(0.6)$ \\
\bottomrule
\end{tabular*}
\end{table*}

Additionally, \textbf{\footnotesize\autoref{tbl2}} presents the evaluation results of different methods on the external dataset, SZC-T1. From the ORG row, it can be observed that the original average misalignment of the external dataset, SZC-T1($\mathit{\textit{DSC}}$$\texttt{:}$$0.748$), is significantly smaller compared to it in the internal dataset, T1dataset210($\mathit{\textit{DSC}}$$\texttt{:}$$0.680$). However, the variations in image acquisition devices have increased the challenges of generalization. Our method achieves the best results, with a Dice Similarity Coefficient ($\mathit{\textit{DSC}}$) of $0.809$, an endocardial Hausdorff Distance ($\mathit{\textit{HD\_endo}}$) of $4.60$, and an epicardial Hausdorff Distance ($\mathit{\textit{HD\_epi}}$) of $4.39$. Compared to VoxelMorph, our method improves the $\mathit{\textit{DSC}}$ by $6.2\%$, reduces the $\mathit{\textit{HD\_endo}}$ by $6.5\%$ in the endocardial region, and reduces the $\mathit{\textit{HD\_epi}}$ by $7.8\%$ in the epicardial region. Furthermore, in terms of the topology preservation performance, our method achieves performance equivalent to SyN and surpasses methods DisQ and VM by a significant margin with a decrease in the number of pixels with non-positive Jacobian values of 99.49\% and 99.53\% respectively. In addition, compared with the contour-based local registration method Cb-Reg, our proposed RS-MOCO performs better in the external dataset. Cb-Reg utilizes manually delineated myocardial contours for contour-based image registration, making its registration results highly dependent on accurately delineated myocardial contours and standardized preprocessing. However, due to limitations in equipment and acquisition technology, the data collected clinically in hospitals is of poor quality. For instance, blurred myocardial contours and low image resolution can lead to errors in the delineation of the myocardial contours and difficulties in performing standardized preprocessing, thereby resulting in registration failure of Cb-Reg. In contrast, our proposed RS-MOCO mainly performs registration on the image by virtue of the powerful abstract feature extraction ability of the neural network, and does not rely on the manually delineated myocardial segmentation labels at the testing stage, so it exhibits stronger generalization ability and convenience.

\begin{table*}[width=1.25\linewidth,cols=6,pos=h]\scriptsize
\captionsetup[table]{labelfont=rm,textfont={rm,rm}}
\rm\caption{\rm Performance comparisons of different registration methods on the SZC-T1 dataset.}\label{tbl2}
\renewcommand\arraystretch{1.5}
\begin{tabular*}{\tblwidth}{@{} cccccc@{} }
\toprule
\textrm{Method} & \textrm{DSC} & $|J_{\Phi}|$$\leq$$0$ & \textrm{HD\_endo(px)} & \textrm{HD\_epi(px)} & \textrm{Time(s)} \\
\midrule
\textrm{ORG} & $0.748(0.123)$ & $-$ & $5.04(1.13)$ & $4.80(1.12)$ & $-$ \\
\textrm{Cb-Reg} & $0.779(0.104)$ & $-$ & $4.66(1.08)$ & $4.60(1.11)$ & $172.2(5.7)$ \\
\textrm{SyN} & $0.792(0.110)$ & \boldmath{$0.509(0.427)$} & $4.75(1.02)$ & $4.56(1.10)$ & $286.1(10.8)$ \\
\textrm{DisQ} & $0.781(0.118)$ & $160.05(131.8)$ & $4.83(1.11)$ & $4.62(1.12)$ & $6.6(0.5)$ \\
\textrm{VM} & $0.762(0.123)$ & $171.46(165.2)$ & $4.92(1.12)$ & $4.76(1.13)$ & $6.8(0.3)$ \\
\textrm{Ours(AffN)} & $0.770(0.118)$ & $-$ & $4.89(1.11)$ & $4.71(1.12)$ & $2.7(0.1)$ \\
\textrm{Ours} & \boldmath{$0.809(0.085)$} & $0.812(0.531)$ & \boldmath{$4.60(1.07)$} & \boldmath{$4.39(1.12)$} & $6.5(0.4)$ \\
\bottomrule
\end{tabular*}
\end{table*}

\begin{figure}  % [width=\textwidth] 和 *表示两栏
	\centering
		\includegraphics[scale=.55]{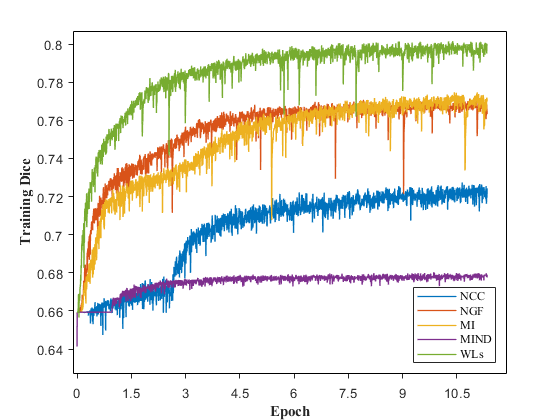}
	\caption{\textrm{The trend plot of dice scores trained with different image similarity losses of the proposed deformable registration network (RegN) before alternative training.}}
	\label{Fig. 11}
\end{figure}

\subsubsection*{\textnormal{\textit{\fontsize{9}{20}\selectfont 4.3.3. Ablation study analysis}}}
The left sub-table in \textbf{\footnotesize\autoref{tbl3}} presents a comparison of results on the T1dataset210 dataset using different image similarity metrics to validate the effectiveness of the proposed weighted image similarity metric, $\mathit{\textit{WLs}}$. The results demonstrate the superiority of $\mathit{\textit{WLs}}$ over all single-used image similarity metrics. In comparison to the suboptimal $\mathit{\textit{MI}}$, the use of $\mathit{\textit{WLs}}$ in RS-MOCO shows a $3.2\%$ increase in $\mathit{\textit{DSC}}$, a $3.7\%$ decrease in $\mathit{\textit{HD\_endo}}$ and a $3.6\%$ decrease in $\mathit{\textit{HD\_epi}}$. This indicates that $\mathit{\textit{WLs}}$ can better capture the significantly varying intensity and structure information in T1w images, thereby addressing the significant contrast variation issue and enhancing the accuracy of registration. Additionally, \textbf{\footnotesize\autoref{Fig. 11}} illustrates the trend of dice variation during network training with different image similarity metrics used in the image similarity loss function. It is evident that $\mathit{\textit{WLs}}$ can accelerate model convergence and significantly improve model performance.

On the other hand, the right sub-table in \textbf{\footnotesize\autoref{tbl3}} presents a comparison of results on the T1dataset210 dataset by omitting various modules to verify their effectiveness. The results demonstrate that the performance of RS-MOCO shows varying degrees of decline when AffN, SegN, or CBAM are excluded. Overall, CBAM enables the model to focus more on crucial information, contributing the most to the improvement in model performance, with a $5.6\%$ increase in $\mathit{\textit{DSC}}$, a $9.6\%$ decrease in $\mathit{\textit{HD\_endo}}$ and a $8.7\%$ decrease in $\mathit{\textit{HD\_epi}}$. However, the guiding registration role of SegN should not be overlooked, as it leads to a $2.2\%$ increase in $\mathit{\textit{DSC}}$, a $5.1\%$ decrease in $\mathit{\textit{HD\_endo}}$ and a $6.0\%$ decrease in $\mathit{\textit{HD\_epi}}$. Although AffN does not have a significant promotion effect on overall results, it contributes to reducing the training time of RegN. This reduction in training time is attributed to AffN's pre-registration function, which largely corrects linear displacements prior to RegN's processing, and does not compromise the final performance of the model. Additionally, the use of the proposed bidirectional consistency and local orientation consistency constraint, BLOC, yields anticipated outcomes by trading small precision losses ($1.3\%$) for a substantial decrease ($99.4\%$) of the average number of voxels with a non-positive Jacobian determinant, thereby facilitating more desirable registration outcomes by effectively suppressing the occurrence of significant folds and abnormal distortions within the deformation field.

\begin{table*}\scriptsize  % * 必须加
\centering
    \centering\caption{\textrm{Ablation study: performance comparisons of different similarity loss metrics (left) and different missing modules (right) on T1dataset210 dataset. "(only)" means single image similarity metric. "AffN(w/o)" means "the network which removes AffN module only", and so on. "Ours(w)" means "our proposed network with all modules".}}
    \label{tbl3}
    \setlength{\tabcolsep}{2pt}
    \renewcommand\arraystretch{1.5}
    \begin{tabular}{p{45pt}p{50pt}p{50pt}p{50pt}} %  p{40pt} 列宽
        \hline   % 标题
        \makecell[c]{\textrm{Metric}} & \makecell[c]{\textrm{DSC}} & \makecell[c]{\textrm{HD\_endo(px)}} & \makecell[c]{\textrm{HD\_epi(px)}} \\
        \hline
        \makecell[c]{\textrm{ORG}} & \makecell[c]{$0.680(0.194)$} & \makecell[c]{$7.80(4.02)$} & \makecell[c]{$8.03(3.79)$} \\
        \makecell[c]{\textrm{NCC(only)}} & \makecell[c]{$0.750(0.166)$} & \makecell[c]{$7.10(2.91)$} & \makecell[c]{$7.45(2.56)$} \\
        \makecell[c]{\textrm{NGF(only)}} & \makecell[c]{$0.799(0.151)$} & \makecell[c]{$6.49(2.53)$} & \makecell[c]{$7.04(2.25)$} \\
        \makecell[c]{\textrm{MI(only)}} & \makecell[c]{$0.801(0.135)$} & \makecell[c]{$6.43(2.37)$} & \makecell[c]{$6.96(2.20)$} \\
        \makecell[c]{\textrm{MIND(only)}} & \makecell[c]{$0.705(0.167)$} & \makecell[c]{$7.68(3.59)$} & \makecell[c]{$7.82(3.46)$} \\
        \makecell[c]{\textrm{WLs}} & \makecell[c]{\boldmath{$0.827(0.093)$}} & \makecell[c]{\boldmath{$6.19(2.06)$}} & \makecell[c]{\boldmath{$6.71(1.95)$}} \\
        \hline
    \end{tabular}
    % \begin{tabular}{p{28pt}p{30pt}p{38pt}p{28pt}}
    \renewcommand\arraystretch{1.5}
    \begin{tabular}{p{50pt}p{50pt}p{55pt}p{50pt}p{50pt}}
        \hline   % 标题
        \makecell[c]{\textrm{Module}} & \makecell[c]{\textrm{DSC}} & \makecell[c]{$|J_{\Phi}|$$\leq$$0$} & \makecell[c]{\textrm{HD\_endo(px)}} & \makecell[c]{\textrm{HD\_epi(px)}} \\
        \hline
       \makecell[c]{\textrm{ORG}} & \makecell[c]{$0.680(0.194)$} & \makecell[c]{$-$} & \makecell[c]{$7.80(4.02)$} & \makecell[c]{$8.03(3.79)$} \\
        \makecell[c]{\textrm{AffN(w/o)}} & \makecell[c]{$0.822(0.119)$} & \makecell[c]{$1.561(1.215)$} & \makecell[c]{$6.23(2.15)$} & \makecell[c]{$6.79(2.11)$} \\
        \makecell[c]{\textrm{SegN(w/o)}} & \makecell[c]{$0.809(0.134)$} & \makecell[c]{\boldmath{$0.818(0.870)$}} & \makecell[c]{$6.52(2.64)$} & \makecell[c]{$7.14(2.32)$}  \\
        \makecell[c]{\textrm{CBAM(w/o)}} & \makecell[c]{$0.783(0.161)$} & \makecell[c]{$1.108(0.804)$} & \makecell[c]{$6.85(2.83)$} & \makecell[c]{$7.35(2.54)$}  \\
        \makecell[c]{\textrm{BLOC(w/o)}} & \makecell[c]{\boldmath{$0.838(0.123)$}} & \makecell[c]{$165.751(161.235)$} & \makecell[c]{$6.28(2.12)$} & \makecell[c]{$6.78(2.07)$}  \\
        \makecell[c]{\textrm{Ours(w)}} & \makecell[c]{$0.827(0.093)$} & \makecell[c]{$0.921(0.552)$} & \makecell[c]{\boldmath{$6.19(2.06)$}} & \makecell[c]{\boldmath{$6.71(1.95)$}} \\
        \hline
    \end{tabular}
\end{table*}
\end{small}

\section{Conclusion}
\begin{small}
This study proposes a deep learning-based and topology-preserving image registration framework, RS-MOCO, for motion correction of cardiac T1 mapping. The proposed deformable registration network achieves topology-preserving registration by the use of the novel BLOC constraint. The BLOC constraint ensures normal and desirable registration results, i.e., negligible discontinuities, folds, or abnormal distortions in the warped images. By incorporating the proposed weighted image similarity metric, WLs, both affine and deformable registration networks address the significant contrast and structure variation issue in T1w images. The affine registration network replaces the conventional affine pre-alignment method implemented on other platforms, making RS-MOCO an end-to-end motion correction framework. Besides, the dual-domain attention module and the additional semi-supervised segmentation network both enhance the registration performance. Interestingly, even imperfect segmentation results can improve the performance of the registration network. In our experiments, there were some discrepancies between the myocardial segmentation predicted by SegN and those manually annotated. However, compared to unsupervised registration learning, these imperfect segmentation results can still guide RegN to better comprehend the anatomical structure of the images to a certain extent, thereby improving the performance of RegN. Extensive comparative experiments and ablation studies demonstrate the effectiveness and superiority of our method.

In future research, building upon RS-MOCO, we intend to directly utilize the myocardial segmentation results from the semi-supervised segmentation network for fully automated quantification analysis of myocardial T1 values. This approach holds greater feasibility compared to general supervised segmentation networks in terms of the following aspects: 1) The semi-supervised segmentation network, SegN, within RS-MOCO, is able to learn by using partial labeled data combined with a large amount of unlabeled data, thereby reducing the dependency on the full set of manual segmentation labels. 2) SegN is expected to achieve segmentation performance equivalent to a fully supervised segmentation network. 3) RS-MOCO can perform both myocardial segmentation and registration tasks simultaneously, which is more convenient than performing the registration and segmentation steps separately in two individual networks.
\end{small}
% \end{linenumbers}

\vspace{8pt}
\noindent\normalsize  \textbf{Declaration of competing interest}

\begin{small}
We declare that we have no potential financial or personal interests with any other individuals or organizations.
\end{small}

\vspace{8pt}
\noindent\normalsize  \textbf{Acknowledgements}

\begin{small}
This study was supported in part by the National Key R\&D Program of China nos. 2021YFF0501402, 2020YFA0712200. National Natural Science Foundation of China under grant No. 62322119, 12226008, 81971611, U21A6005. The Key Laboratory for Magnetic Resonance and Multimodality Imaging of Guangdong Province under grant no. 2020B1212060051, the Guangdong Basic and Applied Basic Research Foundation no. 2021A1515110540; Shenzhen Science and Technology Program under grant No. RCYX20210609104444089, JCYJ20220818101205012.
\end{small}

% \section{Bibliography}
\begin{tiny}
\bibliographystyle{elsarticle-num}
\bibliography{arxiv}
\end{tiny}

\end{document}